\documentclass{article}
\usepackage{spconf,graphicx}
\usepackage{comment}
\usepackage{amsmath,amssymb,mathtools, commath}
\usepackage{multicol}
\usepackage{booktabs}
\usepackage{tabularx}
\usepackage{multirow}
\usepackage{balance}
\usepackage{pgfplots}
\usepackage[noadjust]{cite}
\usepackage[caption=false,font=normalsize,labelfont=sf,textfont=sf]{subfig}
\usepackage{tikz}

\pgfplotsset{compat=newest}

\usetikzlibrary{positioning,spy,backgrounds,calc,arrows}
\graphicspath{{./images/}}

\newcommand{\bs}{\boldsymbol}
\newcommand{\mbf}{\mathbf}

\title{Model Adaptation for image reconstruction using \\Generalized Stein's Unbiased Risk Estimator}

\name{Hemant Kumar Aggarwal and  Mathews Jacob 
	\thanks{This work is supported by 1R01EB019961-01A1 and 1 R01 AG067078-01A1. This work was conducted on an MRI instrument funded by 1S10OD025025-01 }}
\address{University of Iowa, Iowa, USA}%

\begin{document}
%\ninept
%
\maketitle
\begin{abstract}

Deep learning image reconstruction algorithms often suffer from model mismatches when the acquisition scheme differs significantly from the forward model used during training. We introduce a Generalized Stein's Unbiased Risk Estimate (GSURE) loss metric to adapt the network to the measured k-space data and minimize model misfit impact. Unlike current methods that rely on the mean square error in k-space, the proposed metric accounts for noise in the measurements. This makes the approach less vulnerable to overfitting, thus offering improved reconstruction quality compared to schemes that rely on mean-square error. This approach may be useful to rapidly adapt pre-trained models to new acquisition settings (e.g., multi-site) and different contrasts than training data.

\end{abstract}
\section{Introduction}

The reconstruction of images from a few noisy measurements is a central problem in several modalities, including MRI, computer vision, and microscopy. Classical methods, including compressed sensing (CS), pose the recovery as an optimization scheme. The cost function in CS is the sum of a data consistency term involving a numerical forward model of the acquisition scheme and a regularization term that exploits image priors~\cite{lustig2008compressed,candes2007sparsity}. 

Recently, deep learning algorithms are emerging as powerful alternatives offering improved performance over CS-based methods that often rely on carefully handcrafted regularization priors. Most deep learning methods for image reconstruction rely on learning of trainable convolutional neural network (CNN) modules within the network using fully sampled training images~\cite{modl,casecadeDynamic,jongCT}. In addition to computational efficiency, these deep learning based methods provide improved image quality than classical CS-based approaches.

Unlike CS priors that only depend on the image, the learned CNN modules often depend on the specific forward model used in training. In many cases, the actual acquisition model can differ significantly from those used to train the network. In those cases, deep learning methods may offer sub-optimal image quality. In the MR imaging context, several factors can contribute to the above model mismatches, including differences in acceleration factors, sampling patterns, the amount of measurement noise, specific parallel MRI coils, inter-site variability, inter-scanner variability, as well as differences in image content and contrast.

Practitioners rely on training the network with several forward models to minimize model mismatch related dependence~\cite{modl,sigmanet,gan_cyclic,dagan,casecadeDynamic}. However, even these models have some sensitivity to model mismatch. It is often not practical to train the network to each setting because of the lack of fully-sampled training data corresponding to every case. To minimize this challenge, several authors have proposed to fine-tune the pre-trained networks using the error between the actual measurements and the ones made on the recovered images~\cite{dip2018}. A challenge with this scheme is the need for careful early stopping. Specifically, CNNs often have sufficient capacity to learn measurement noise. Therefore,  without early stopping, the algorithm can overfit the few measurements, resulting in degraded performance. Some authors have proposed to add additional priors to restrict the network parameters to not deviate significantly from the original ones~\cite{sigmanet}. Loss functions that only use part of the data were introduced~\cite{ssduft}.

We introduce a loss function based on Stein's unbiased risk estimator (SURE)~\cite{sure} to adapt a pre-trained deep image reconstruction network to a new acquisition scheme and image content. Unlike prior approaches~\cite{sigmanet,dip2018,ssduft} that do not account for measurement noise, the proposed approach accounts for the noise statistics and is less vulnerable to overfitting. The proposed model adaptation scheme will work with both model-based algorithms~\cite{modl,hammernik} as well as direct-inversion methods \cite{jong2019kspace,ronneberger2015unet}. 

Stein's unbiased risk estimator (SURE)~\cite{sure} is an unbiased estimator for mean-square-error (MSE). LDAMP-SURE~\cite{metzler2018} utilizes this SURE estimate to train CNN denoisers in an unsupervised fashion. LDAM-SURE also proposes to train denoisers within an unrolled network in a layer-by-layer manner for image recovery from undersampled measurements~\cite{metzler2018}. The GSURE approach~\cite{eldarGSURE} extends SURE to inverse problems and considers an unbiased estimate of the MSE in the range space of the measurement operator. A challenge in using GSURE~\cite{eldarGSURE} to train deep networks for inverse problems is the poor approximation of the actual MSE by the projected MSE, especially when the range space is small~\cite{metzler2018}. We recently developed an ENsembled SURE (ENSURE)~\cite{ensure} approach for unsupervised learning to overcome this problem. We showed that an ensemble of sampling patterns can well approximate the projected MSE as weighted MSE.

In this work, we use the GSURE~\cite{eldarGSURE} approach to adapt a pre-trained network to a new acquisition setting, only using the undersampled measurements. Unlike~\cite{ensure}, where a network is trained from scratch,  we consider adopting a pre-trained network only using the undersampled measurements of a single image. Our results show that the GSURE-based model adaptation (GSURE-MA) offers improved performance even in highly undersampled settings compared to existing approaches~\cite{ssduft, dip2018}.

\section{Proposed Method}
The image acquisition model to acquire the noisy and undersampled measurements $\bs y \in \mathbb C^n$ of an image $\bs x \in \mathbb C^m$ using the forward operator $\mathcal A$ can be represented as
\begin{equation}
	\label{eq:fwd}
	\bs y=\mathcal A \bs x +\bs n
\end{equation}
Here, we assume that noise $\bs n$ is Gaussian distributed with mean zero and standard deviation $\mbf \sigma$ such that $\bs{n}\sim \bs N(0,\mbf \sigma)$. Define regridding reconstruction as {$\bs u=\mathcal A^H \bs y$} that lives in a subspace of $\mathbb C^m$, specified by $\mathcal V$.  The recovery using a deep neural network $f_\Phi$ with trainable parameters $\Phi$ can be represented as
\begin{equation}
	\label{recon}
	\bs{\widehat x} = f_{\Phi}(\bs {u}). 
\end{equation}
Here $f_\Phi$ can be a direct-inversion or a model-based deep neural network. Supervised deep learning methods compare the recovered image $\widehat{\bs x}$ with fully sampled ground truth image $\bs x$ using 
\begin{equation}
	\label{eq:mse}
	\text{MSE}= \mathbb E_{\bs x \sim \mathcal M}~ \| \bs{\widehat x} -  \bs x \|_2^2
\end{equation}
 to train the reconstruct network.

\begin{figure}	\centering
	\subfloat[data-term] {	
		\includegraphics[ width=.8\linewidth]{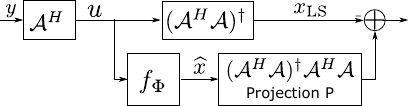}	
	}
	
	\subfloat[divergence-term] {	
		\includegraphics[width=.7\linewidth]{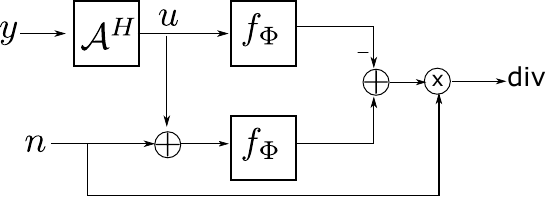}	\vspace{-2em}
	}
	\caption{\footnotesize{The implementation details of the GSURE based loss function for model adaptation. (a) shows the calculation of data-term. (b) shows the calculation of the divergence term. Here we pass the regridding reconstruction and its noisy version through the network and find the error between the two terms. Then we take the inner product between this error term and the noise to get an estimate of the network divergence divergence.}}
	\label{fig:gsure} \vspace{-2em}
\end{figure}

The deep network $f_{\Phi}$ is often sensitive to the specific forward model $\mathcal A$ in \eqref{eq:fwd} and the class of images $\mathcal M$. As discussed before, the above trained networks are vulnerable to model mismatches, when the acquisition scheme or the type of images are different. We hence consider the adaptation of the trained network $f_{\Phi}$ to the specific images based on the available measurements, assuming that fully sampled ground truth to perform training using MSE, as in~\eqref{eq:mse}, is not available. 

\begin{figure}[b!]	\centering
	\subfloat[\scriptsize{Training Mask M$_0$}] {	
		\includegraphics[ width=.28\linewidth]{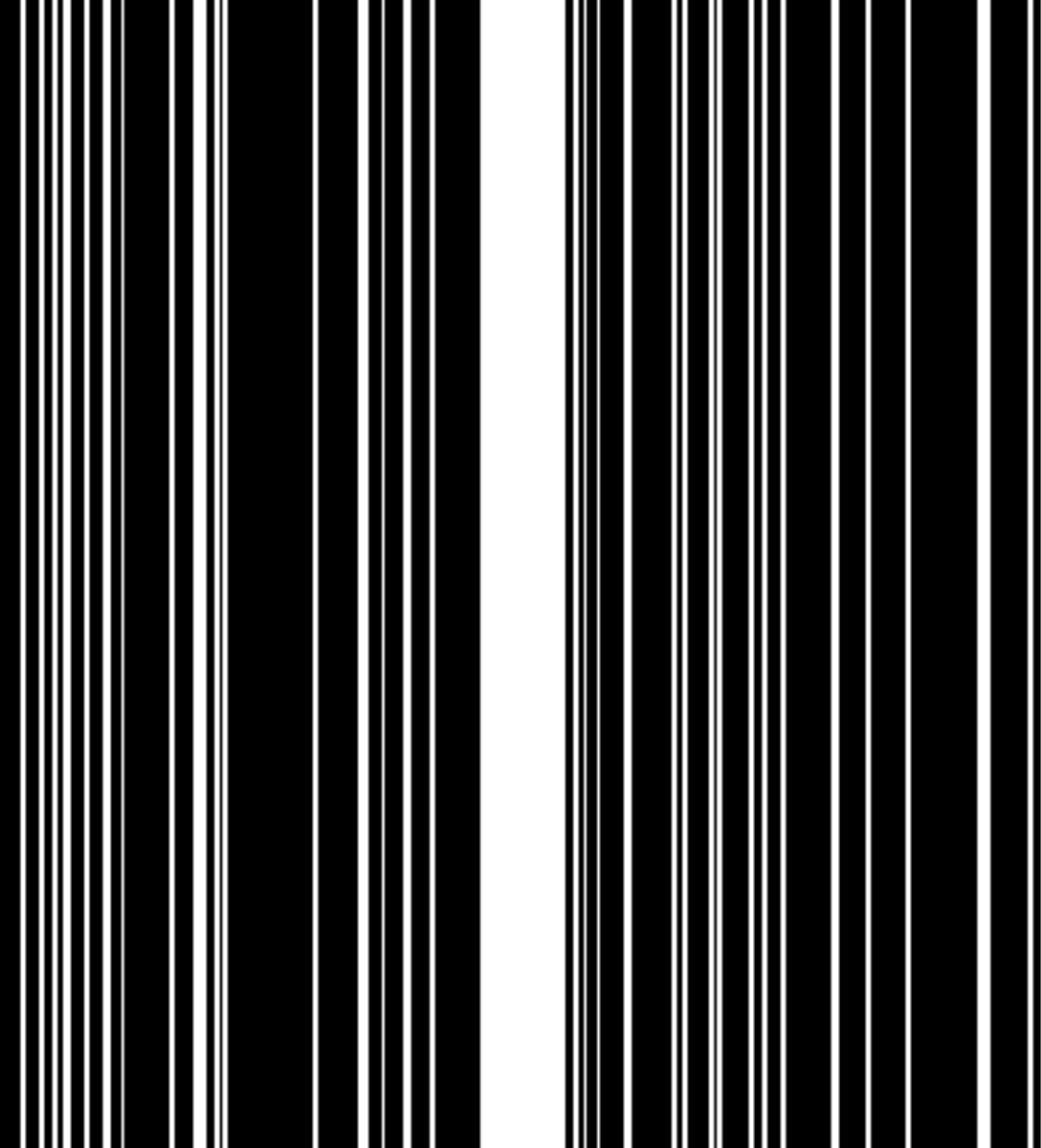}	
	} 	\qquad
	\subfloat[\scriptsize{Testing Mask M$_1$}] {	
		\includegraphics[width=.28\linewidth]{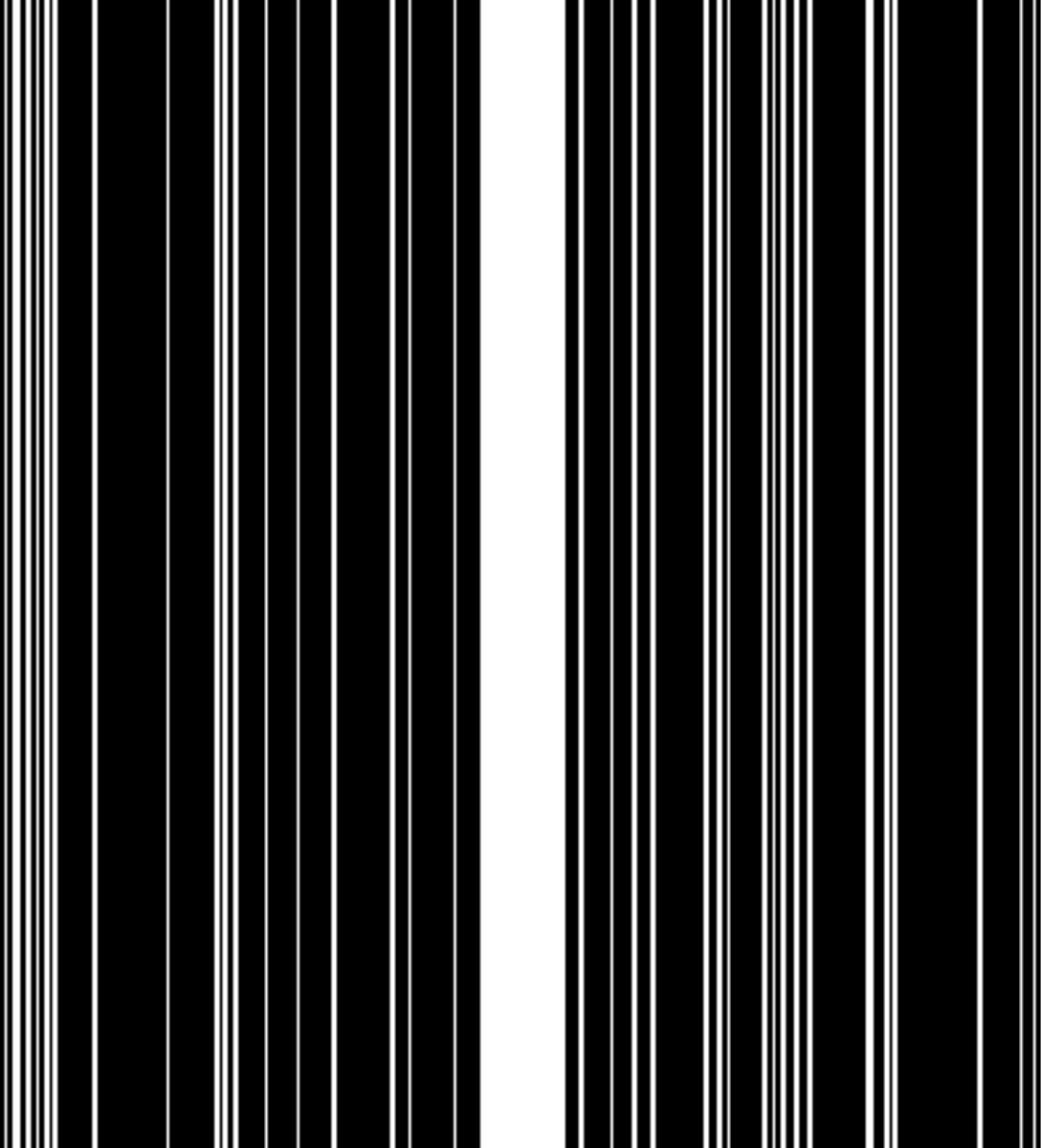}	
	}
	\caption{\footnotesize{The 1D Cartesian sampling masks used for training and testing for the first experiment.}}
	\label{fig:masks1d}
\end{figure}

A simple approach is to perform model adaptation using the available noisy measurements only with the loss function:
\begin{equation}
	\label{eq:dip}
	\text{DIP-MA}= \mathbb E \| \mathcal A~f_{\Phi}(\bs u) -  \bs y \|_2^2,
\end{equation}
where $\Phi$ is initialized with the parameters of the trained model. Due to the similarity of this approach to \cite{dip2018}, we term this approach as model adaptation using deep image prior (DIP-MA). Because the measurements $\bs y$ are noisy, DIP-MA is vulnerable to overfitting. Early termination and the use of additional regularization priors to restrict the deviation of $\Phi$ from the pretrained ones are used \cite{sigmanet}. 

We propose to use GSURE~\cite{eldarGSURE} loss function that explicitly accounts for the noise in the measurements to minimize overfitting issues. We denote the projection to this subspace as $\mbf P = (\mathcal A^H \mathcal A)^{\dag}\mathcal A^H\mathcal A$, where $\dag$ denotes the pseudo-inverse. The GSURE approach is an unbiased estimate for the projected MSE, denoted by $\|\mathbf P(\bs{\widehat x} - \bs x)\|^2$: 
\begin{equation} \label{gsure}
	\mathcal L =~ \underbrace{\mathbb E_{\bs u} \left [ \|\mbf P \bs{\widehat x} -\bs x_{\text{LS}}  \|_2^2 \right]} _{\mathrm{data~ term}} ~+
	 ~\underbrace{2 \mathbb E_{\bs{u}} \left [\nabla_{\bs{u}} \cdot  f_\Phi(\bs{u})  \right ]}_ {\mathrm{divergence}}. 
\end{equation}
Here $\bs x_{\text{LS}}=(\mathcal A^H \mathcal A)^{\dag} \bs u$ is a least-square estimate. The second term is a measure of the divergence of the network and is computed using the Monte-Carlo approach \cite{mcsure}. This term acts as a network regularization term, this minimizing the risk of overfitting. Fig.~\ref{fig:gsure} shows the implementation details of data-term and the divergence term.

\section{Experiments and Results}
\begin{figure*}
	\input{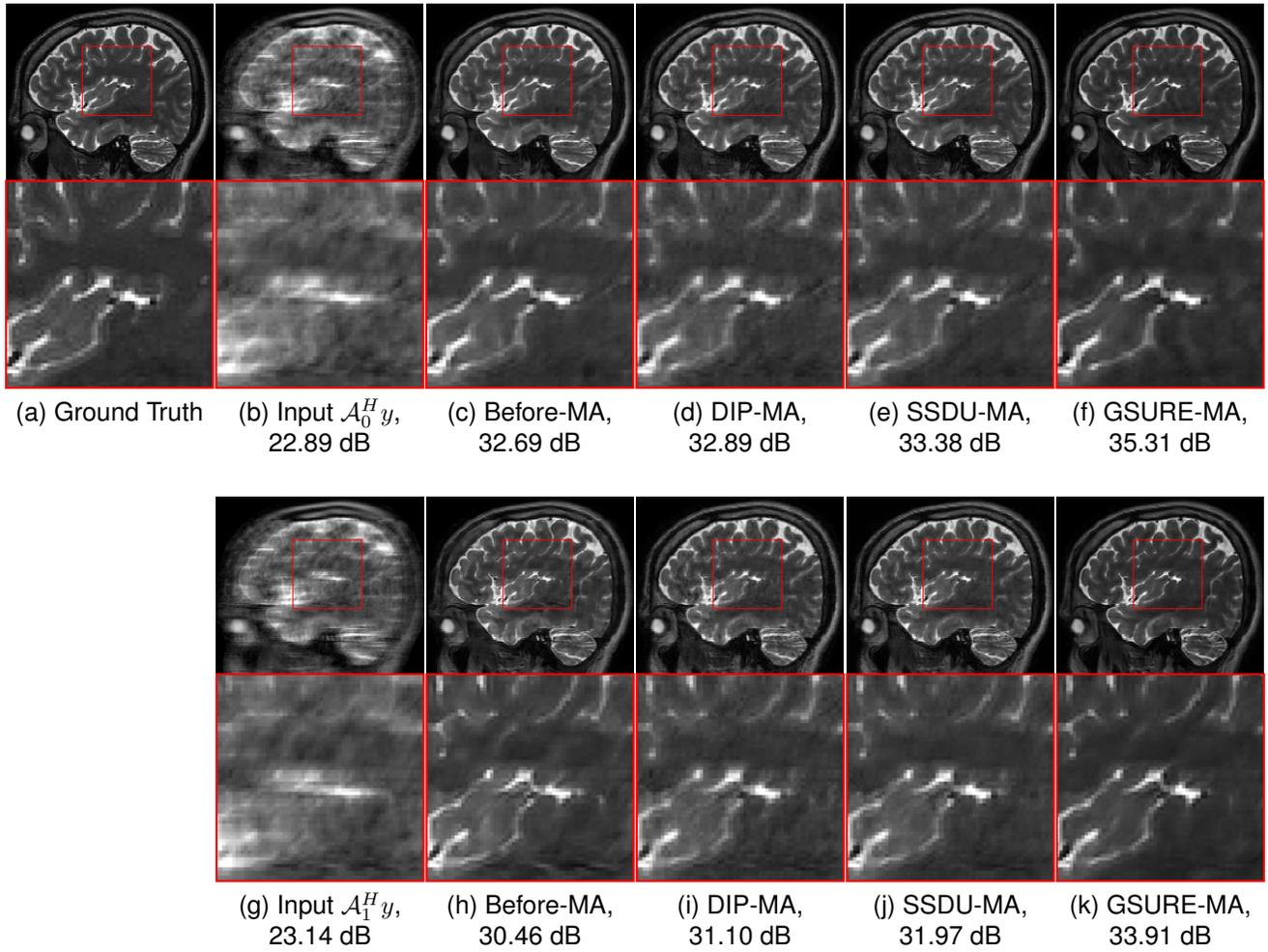}
	\caption{Experimental results for the 1D Cartesian sampling mask on a testing slice. The training and testing forward model $\mathcal A_0$ and $\mathcal A_1$ when applied on a ground truth image (a) from the testing data lead to the re-gridding reconstruction $\mathcal A_0^Hy$ (b) and $\mathcal A_1^Hy$ (g), respectively.  (b-f) shows testing results on the forward model $\mathcal A_0$ that was used during training. (g-k) shows results with a different forward model $\mathcal A_1$, not seen during training. Red box shows a zoomed portion of the images. }
	\label{fig:results1d}
\end{figure*}
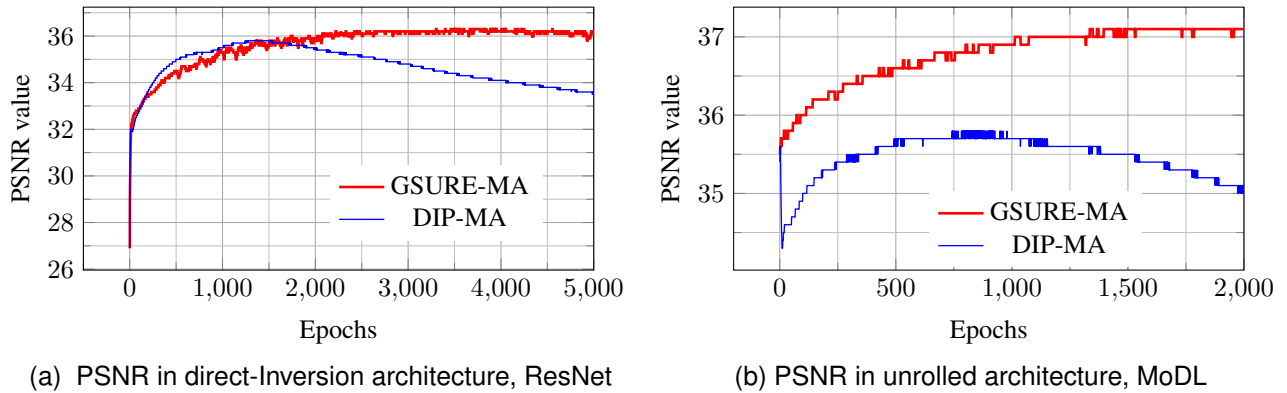
\begin{figure*} 
	\centering
	\subfloat[ PSNR in direct-Inversion architecture, ResNet]{ 
		\centering
\begin{tikzpicture}
 \begin{axis}[
   width=.47\linewidth,
   height=2.in,
   xlabel=Epochs,
   ytick={26,28,...,37},
   xmax=5000,
   ylabel=PSNR value,
   grid=both,   
   grid style={line width=.1pt, draw=gray!50},
   major grid style={line width=.2pt,draw=gray!70},
   minor tick num=1,
   legend style={at={(.9,0.4)},draw=none},
   ]
   \addplot[red,line  width=1]  table [x=idx, y=ensure, col sep=comma] {psnrvaluesdw.csv};
   \addlegendentry{GSURE-MA};
   \addplot[mark=.,blue,line   width=.5]  table [x=idx, y=mse, col sep=comma] {psnrvaluesdw.csv};
   \addlegendentry{DIP-MA};
 \end{axis}
\end{tikzpicture}
		\label{fig:resnetPlot}		
	}
	\subfloat[PSNR in unrolled architecture, MoDL]{
		\centering
\begin{tikzpicture}
 \begin{axis}[
   width=.47\linewidth,
   height=2.in,
   xlabel=Epochs,
   xmax=2000,
   ytick={33,34,35,36,37,38},
   ylabel=PSNR value,   
   grid=both,   
   grid style={line width=.1pt, draw=gray!50},
   major grid style={line width=.2pt,draw=gray!70},
   minor tick num=1,
   legend style={at={(.8,0.3)},draw=none},
   ]
   \addplot[red,line  width=1]  table [x=idx, y=ensure, col sep=comma] {psnrvaluesdc.csv};
   \addlegendentry{GSURE-MA};
   \addplot[mark=.,blue,line   width=.5]  table [x=idx, y=mse, col sep=comma] {psnrvaluesdc.csv};
   \addlegendentry{DIP-MA};
 \end{axis}
\end{tikzpicture}
		\label{fig:modlPlot}		
	}
	\caption{\footnotesize{These plots show the variation in PSNR values with the model adaptation epochs on a single test image using DIP-MA and proposed GSURE-MA strategies. The ResNet and MoDL architectures were fine-tuned for 5000 and 2000 epochs, respectively.}}
	\label{fig:plots}
\end{figure*}

We consider a publicly available~\cite{modl} parallel MRI brain data obtained using 3T GE MR750w scanner  at the University of Iowa. The matrix dimensions were $256\times256\times208$ with a 1~mm isotropic resolution. Fully sampled multi-channel brain images of nine volunteers were collected, out of which data from five subjects were used for training. The data from two subjects were used for testing and the remaining two for validation.

We evaluate the performance of the proposed model-adaption technique in both the direct-inversion-based networks and unrolled model-based networks. Specifically, we use ResNet18 as the direct-inversion network and the MoDL architecture as the unrolled network. The ResNet18 has $3\times3$ convolution filters and 64 feature maps at each layer. The real and imaginary components of complex data were used as channels in all the experiments. For the MoDL architecture, we use three unrolling steps, each having a ResNet18 followed by a data-consistency step. The network weights are shared over the three unrolls.

We compare the proposed GSURE-MA approach with DIP-MA and self-supervised learning via deep undersampling (SSDU)~\cite{ssduft}. For model-adaption using SSDU (SSDU-MA), we utilized 60\% of the measured k-space data for the data-consistency and the remaining 40\% for the loss-function, as suggested in SSDU-MA~\cite{ssduft}.  

The first experiment demonstrates the benefits of model-adaptation for 1D multichannel Cartesian sampling. Fig.~\ref{fig:masks1d} shows the training mask $M_0$ and testing mask $M_1$, corresponding to training and testing forward models $\mathcal A_0$ and  $\mathcal A_1$ respectively.  We first performed a  supervised training of the MoDL architecture assuming $\mathcal A_0$ on 360 training slices. After training, we tested the performance of the learned model on 100 test slices from a different subject using forward models $\mathcal A_0$ as well as $\mathcal A_1$. 

Fig.~\ref{fig:results1d} shows both qualitative and quantitative results on models $\mathcal A_0$ and $\mathcal A_1$. 
%Fig.~\ref{fig:results1d}(b-f) shows the experiments when the model $\mathcal A_0$ is used, while Fig.~\ref{fig:results1d}(h-k) corresponds to $\mathcal A_1$. 
Fig.~\ref{fig:results1d}(h) shows that the MoDL architecture is relatively robust to the change in the forward model. The DIP-MA scheme offers relatively modest improvement, which are outperformed by SSDU-MA. It is evident from PSNR values as well as from visual comparisons that the proposed GSURE-MA leads to the best quality as compared to existing approaches. Specifically, accounting for the noise during the model adaptation phase results in improved performance. We note that the GSURE-MA scheme offers improved performance even when $\mathcal A_0$ is used. We attribute this to the differences in image content, compared to the ones used for training. 

\begin{table}
	\caption{\footnotesize{Table shows PSNR (dB) values of the reconstructed test dataset at four different acceleration (Acc.) factors ranging from two-fold (2x)  to eight-fold (8x) acceleration. The pre-training was performed with the 6x acceleration setting.  }}
	\label{tab:result}	
	\centering
	\begin{tabular}{l|l|cccc} \toprule
		&Acc.                                      & 2x    &  4x  &  \textbf{6x}   &  8x   \\ \midrule
&Input, $\mathcal A^T b$	               & 30.23 & 24.80 & \textbf{22.96}& 22.27 \\ \midrule
   \multirow{3}{1.22cm}{Dir.~Inv. ResNet}&Before-MA & 23.37 & 29.37 & \textbf{32.10}& 30.34 \\
		&DIP-MA	                                   & 33.17 & 34.06 & \textbf{33.21}& 32.28 \\
		&GSURE-MA                                  & 35.16 & 35.79 & \textbf{34.86}& 33.66 \\ \midrule 
	\multirow{4}{1.22cm}{Unrolled MoDL}	&Before-MA & 28.37 & 35.10 & \textbf{35.35}& 33.99 \\
		&DIP-MA	                                   & 37.72 & 33.50 & \textbf{31.94}& 31.15 \\
    	&SSDU-MA	                               & 34.46 & 33.31 & \textbf{30.92} & 29.69 \\
		&GSURE-MA                                  & 39.96 & 37.80 & \textbf{36.08} & 34.97 \\ \bottomrule		 
	\end{tabular}

\end{table}

The graphs in Fig.~\ref{fig:plots} shows a comparison of the DIP based and GSURE based model adaption techniques. The DIP based approach is dependent on the number of epochs. The performance starts dropping after a few iterations and thus DIP-MA requires to manually find the optimal number of iterations. We also observe that GSURE-MA is more stable than DIP-MA and does not require early termination. This behavior is primarily due to the network divergence term that acts as a regularization term in the loss function. Further, we observe from Fig.~\ref{fig:plots}(b), that in the case of unrolled architecture, the maximum PSNR value achieved with GSURE-MA is higher than DIP-MA. Additionally, we note that model-adaptation in unrolled architecture leads to higher PSNR values than the direct-inversion-based approach.

The next experiment demonstrates the model adaptation capabilities of the proposed GSURE-MA method for different acceleration factors. In particular, we train a model for the six-fold (6x) acceleration factor with different 2D random variable density sampling masks. During testing, we evaluate this trained model at 2x, 4x, 6x, and 8x acceleration factor for both direct-inversion and model-based unrolled networks.  Table~\ref{tab:result} summarizes the experimental results of this experiment. SSDU-MA strategy is developed only for unrolled architecture, therefore, its results are calculated for that setting only. Table~\ref{tab:result} shows that the performance of a model trained for 6x acceleration does not work well for 2x acceleration. The rows corresponding to before model adaptation ( Before-MA ) shows the PSNR values of the reconstructed images from the 6x trained model. The proposed GSURE-MA strategy improves the PSNR from 23.37 dB to 35.16 dB in the direct-inversion network and 28.37 dB to 39.96 dB for the unrolled network. Similarly, we see that model adaption improves the reconstruction results for all the accelerations.

\section{Conclusions}
This work proposed a model adaptation strategy to fine-tune a previously trained, deep learned model to the new acquisition operator. We use the GSURE loss function to rapidly adapt a pre-trained model to new acquisition models without the risk of overfitting. We show the preliminary utility of the proposed GSURE-MA scheme for MR image reconstruction.

\balance

%\bibliography{IEEEabrv,bibTexSamp}

\clearpage
\begin{thebibliography}{10}

\bibitem{lustig2008compressed}
Michael Lustig, David~L Donoho, Juan~M Santos, and John~M Pauly,
\newblock ``Compressed sensing {MRI},''
\newblock {\em IEEE signal processing magazine}, vol. 25, no. 2, pp. 72, 2008.

\bibitem{candes2007sparsity}
Emmanuel Candes and Justin Romberg,
\newblock ``Sparsity and incoherence in compressive sampling,''
\newblock {\em Inverse problems}, vol. 23, no. 3, pp. 969, 2007.

\bibitem{modl}
Hemant~K Aggarwal, Merry~P Mani, and Mathews Jacob,
\newblock ``{MoDL}: Model-based deep learning architecture for inverse
  problems,''
\newblock {\em {IEEE} Trans. Med. Imag.}, vol. 38, no. 2, pp. 394--405, 2019.

\bibitem{casecadeDynamic}
Jo~Schlemper, Jose Caballero, Joseph~V Hajnal, Anthony~N Price, and Daniel
  Rueckert,
\newblock ``A deep cascade of convolutional neural networks for dynamic {MR}
  image reconstruction,''
\newblock {\em {IEEE} Trans. Med. Imag.}, vol. 37, no. 2, pp. 491--503, 2018.

\bibitem{jongCT}
Eunhee Kang, Junhong Min, and Jong~Chul Ye,
\newblock ``A deep convolutional neural network using directional wavelets for
  low-dose {X-ray CT} reconstruction,''
\newblock {\em Medical Physics}, vol. 44, no. 10, pp. e360--e375, 2017.

\bibitem{sigmanet}
Kerstin Hammernik, Jo~Schlemper, Chen Qin, Jinming Duan, Ronald~M. Summers, and
  Daniel Rueckert,
\newblock ``{Sigma-Net}: Systematic evaluation of iterative deep neural
  networks for fast parallel {MR} image reconstruction,''
\newblock {\em arXiv preprint arXiv:1912.09278}, 2019.

\bibitem{gan_cyclic}
Tran~Minh Quan, Thanh Nguyen-Duc, and Won-Ki Jeong,
\newblock ``Compressed sensing {MRI} reconstruction using a generative
  adversarial network with a cyclic loss,''
\newblock {\em {IEEE} Trans. Med. Imag.}, vol. 37, no. 6, pp. 1488--1497, 2018.

\bibitem{dagan}
Guang Yang, Simiao Yu, Hao Dong, Greg Slabaugh, Pier~Luigi Dragotti, Xujiong
  Ye, Fangde Liu, Simon Arridge, Jennifer Keegan, Yike Guo, et~al.,
\newblock ``{DAGAN}: Deep de-aliasing generative adversarial networks for fast
  compressed sensing {MRI} reconstruction,''
\newblock {\em {IEEE} Trans. Med. Imag.}, vol. 37, no. 6, pp. 1310--1321, 2017.

\bibitem{dip2018}
Dmitry Ulyanov, Andrea Vedaldi, and Victor Lempitsky,
\newblock ``Deep image prior,''
\newblock in {\em Proceedings of the IEEE Conference on Computer Vision and
  Pattern Recognition}, 2018, pp. 9446--9454.

\bibitem{ssduft}
S.~A. {Hossein Hosseini}, B.~{Yaman}, S.~{Moeller}, and M.~{Ak{\c{c}}akaya},
\newblock ``High-fidelity accelerated mri reconstruction by scan-specific
  fine-tuning of physics-based neural networks,''
\newblock in {\em 2020 42nd Annual International Conference of the IEEE
  Engineering in Medicine Biology Society (EMBC)}, 2020, pp. 1481--1484.

\bibitem{sure}
Charles~M Stein,
\newblock ``Estimation of the mean of a multivariate normal distribution,''
\newblock {\em The annals of Statistics}, pp. 1135--1151, 1981.

\bibitem{hammernik}
Kerstin Hammernik, Teresa Klatzer, Erich Kobler, Michael~P. Recht, Daniel~K.
  Sodickson, Thomas Pock, and Florian Knoll,
\newblock ``{Learning a Variational Network for Reconstruction of Accelerated
  {MRI} Data},''
\newblock {\em Magnetic resonance in Medicine}, vol. 79, no. 6, pp. 3055--3071,
  2017.

\bibitem{jong2019kspace}
Yoseob Han, Leonard Sunwoo, and Jong~Chul Ye,
\newblock ``k-space deep learning for accelerated {MRI},''
\newblock {\em {IEEE} Trans. Med. Imag.}, 2019.

\bibitem{ronneberger2015unet}
Olaf Ronneberger, Philipp Fischer, and Thomas Brox,
\newblock ``U-net: Convolutional networks for biomedical image segmentation,''
\newblock in {\em International Conference on Medical Image Computing and
  Computer-Assisted Intervention (MICCAI)}. Springer, 2015, pp. 234--241.

\bibitem{metzler2018}
Christopher~A Metzler, Ali Mousavi, Reinhard Heckel, and Richard~G Baraniuk,
\newblock ``Unsupervised learning with stein's unbiased risk estimator,''
\newblock {\em arXiv preprint arXiv:1805.10531}, 2018.

\bibitem{eldarGSURE}
Yonina~C Eldar,
\newblock ``Generalized sure for exponential families: Applications to
  regularization,''
\newblock {\em IEEE Transactions on Signal Processing}, vol. 57, no. 2, pp.
  471--481, 2008.

\bibitem{ensure}
Hemant Kumar~A Aggarwal, Aniket Pramanik, and Mathews Jacob,
\newblock ``{ENSURE}: Ensemble stein's unbiased risk estimator for unsupervised
  learning,''
\newblock {\em arXiv:2010.10631}, 2018,
\newblock https://arxiv.org/abs/2010.10631.

\bibitem{mcsure}
Sathish Ramani, Thierry Blu, and Michael Unser,
\newblock ``Monte-carlo sure: A black-box optimization of regularization
  parameters for general denoising algorithms,''
\newblock {\em IEEE Transactions on image processing}, vol. 17, no. 9, pp.
  1540--1554, 2008.

\end{thebibliography}

\end{document}